\documentclass[lettersize,journal]{IEEEtran}
\usepackage{amsmath,amsfonts}
\usepackage{algorithmic}
\usepackage{algorithm}
\usepackage{array}
\usepackage{textcomp}
\usepackage{stfloats}
\usepackage{url}
\usepackage{verbatim}
\usepackage{graphicx}
\usepackage{cite}

\usepackage{amsmath}
\usepackage{amssymb}
\usepackage{bm}
\usepackage{multirow}
\usepackage{kotex}
\usepackage{makecell}

\makeatletter
\newcommand{\thickhline}{%
    \noalign {\ifnum 0=`}\fi \hrule height 1pt
    \futurelet \reserved@a \@xhline
}
\makeatother

\ifCLASSOPTIONcompsoc
  \usepackage[caption=false,font=footnotesize,labelfont=sf,textfont=sf]{subfig}
\else
  \usepackage[caption=false,font=footnotesize]{subfig}
\fi

\begin{document}

\title{A Real-Time Predictive Pedestrian Collision Warning Service for Cooperative Intelligent Transportation Systems Using 3D Pose Estimation}

\author{Ue-Hwan~Kim\textsuperscript{1}*, Dongho Ka\textsuperscript{2}*, Hwasoo Yeo\textsuperscript{2} and~Jong-Hwan~Kim\textsuperscript{3},~\IEEEmembership{Fellow,~IEEE}
\thanks{This work was supported by Institute for Information \& communications Technology Promotion (IITP) grant funded by the Korea government (MSIT) (No.2020-0-00440, Development of Artificial Intelligence Technology that Continuously Improves Itself as the Situation Changes in the Real World).}
\thanks{\textsuperscript{1}Ue-Hwan Kim is with the AI Graduate School, GIST (Gwang-ju Institute of Science and Technology), Gwang-ju, 61005, Repulic of Korea (e-mail: uehwan@gist.ac.kr).}
\thanks{\textsuperscript{2}Dongho Ka and Hwasoo Yeo are with the Department of Civil and Environmental Engineering, KAIST, Daejeon, 34141, Republic of Korea (e-mail: \{kdh910121, hwasoo\}@kaist.ac.kr).}
\thanks{\textsuperscript{3}Jong-Hwan Kim is with the School of Electrical Engineering, KAIST, Daejeon, 34141, Republic of Korea (e-mail: {johkim@rit.kaist.ac.kr}).}
\thanks{*These authors contributed equally to this work.}
}

\markboth{IEEE Transactions on Systems, Man, and Cybernetics: Systems}%
{Shell \MakeLowercase{\textit{et al.}}: A Sample Article Using IEEEtran.cls for IEEE Journals}


\noindent \textbf{Disclaimer.} This work has been submitted to the IEEE for possible publication. Copyright may be transferred without notice, after which this version may no longer be accessible.

\newpage

\maketitle

\begin{abstract}
Minimizing traffic accidents between vehicles and pedestrians is one of the primary research goals in intelligent transportation systems. To achieve the goal, pedestrian orientation recognition and prediction of pedestrian's crossing or not-crossing intention play a central role. Contemporary approaches do not guarantee satisfactory performance due to limited field-of-view, lack of generalization, and high computational complexity. To overcome these limitations, we propose a real-time predictive pedestrian collision warning service (P2CWS) for two tasks: pedestrian orientation recognition ($100.53$ FPS) and intention prediction ($35.76$ FPS). Our framework obtains satisfying generalization over multiple sites because of the proposed site-independent features. At the center of the feature extraction lies 3D pose estimation. The 3D pose analysis enables robust and accurate recognition of pedestrian orientations and prediction of intentions over multiple sites. The proposed vision framework realizes $89.3$\% accuracy in the behavior recognition task on the TUD dataset without any training process and $91.28$\% accuracy in intention prediction on our dataset achieving new state-of-the-art performance. To contribute to the corresponding research community, we make our source codes public which are available at \url{https://github.com/Uehwan/VisionForPedestrian}
\end{abstract}


\section{Introduction}
%
%
%
%
\IEEEPARstart{A}{dvances} in the technology of autonomous driving and advanced driving assistant systems (ADAS) would transform the way the current transportation system works and integrate into people’s daily lives in the near future \cite{neogi2020context, wang2020visual, xu2018reinforcement}. An ideal transportation system enhances the transportation convenience for both drivers and pedestrians but will put more effort into improving safety. Mainly, the system will focus on minimizing traffic accidents between vehicles and pedestrians since the traffic accidents between them could often result in fatalities \cite{goldhammer2019intentions}. In preventing vehicle and pedestrian accidents and securing safety, pedestrian orientation recognition and crossing intention prediction play a vital role as emergency braking 0.16 second in advance could reduce the severity of accident injuries down to 50\% \cite{zhang2020pedestrian, li2016group}.

However, contemporary warning service preventing collision between pedestrian and vehicle possess a few limitations. First, the existing in-vehicle sensor-based methods may overlook upcoming collision risks due to the limited field-of-view (FoV) and distance range. In-vehicle sensors such as Radar and Lidar allow the detection of pedestrians around vehicles, which restricts the effectiveness of collision-warning. Even in-vehicle sensors recognize possible collisions, they would not secure enough time space to handle the risks.

Next, conventional cooperative-intelligent transportation systems (C-ITS), an infrastructure equipped with sensors in various areas such as roads, power poles and traffic light poles, hardly guarantee a real-time operation and generality over multiple sites. Such algorithms integrate multiple deep-learning modules for detection and other data processing steps; the resulting complicated software architectures slow down the overall computation time. Moreover, pevious research groups have developed algorithms specific to their study sites \cite{fang2019intention, saleh2019contextual}; their methods require retraining in new sites for deployment.


To overcome the limitations mentioned above, we propose a real-time predictive pedestrian collision warning service (P2CWS) for C-ITS using 3D pose estimation. The proposed P2CWS utilizes the existing sensors at intersections, which do not entail visibility obstruction, to recognize pedestrian orientations and predict corssing-or-not-crossing intention. Therefore, the proposed P2CWS does not suffer from limited FoV. Moreover, we propose to take advantage of 3D pose estimation in designing P2CWS. The pose analysis with 3D pose estimation becomes more accurate than that of 2D pose estimation since 3D pose estimation employs a 3D human body model as a knowledge base and temporal context within videos. Subsequently, 3D pose estimation allows precise analysis of pedestrian body orientation as well as prediction of pedestrian intention. The proposed pedestrian analysis based on 3D pose estimation simplifies the data processing process and achieves real-time operation; the usage of the generic 3D pose features enables generalization over multiple sites. 

Specifically, we deduce three categories of information for pedestrian intention prediction: 1) pedestrian features, 2) vehicle-to- pedestrian (V2P) interactions, and 3) environmental contexts. The pedestrian features represent the characteristics of the pedestrian of interest and consist of orientation features, the group size, and the speed of the pedestrian. Next, the V2P interactions describe the effect of the nearby vehicle on the pedestrian’s decision-making and include the distance and angle between them and the vehicle’s speed. Moreover, the environmental contexts illustrate the contextual information and comprise crosswalk distance, angle, the pedestrian’s location semantics. Furthermore, we approximate the physical dimensions, i.e., the distance between objects, utilizing a knowledge-base of average object heights. 2D images do not contain full information for physical dimensions; thus, it is difficult to reconstruct the 3D dimensions. We propose to resolve the 3D dimension reconstruction with the knowledge-base. Finally, we predict the pedestrian’s crossing or not-crossing intention utilizing the features extracted.

In summary, the contributions of our work are as follows:
\begin{enumerate}
    \item \textbf{P2CWS Framework Utilizing Vision Sensors at Intersections} : We propose a collision warning framework which could function at multiple sites and guarantee real-time operation.
    \item \textbf{Pedestrian Crossing Intention Prediction} : We propose strategy is not to detect crossing pedestrian only, but to recognize crossing intention considering information such as pedestrian features, vehicle to pedestrian (V2P) interaction, environmental contexts. 
    \item \textbf{Real-Time Operation and Verification at Multiple Sites}: The proposed vision framework guarantees a real-time operation on a modern processing unit ($>$ 30 FPS) and we verify the performance and the universality of the proposed vision framework at multiple study sites.
    \item \textbf{Open Source}: We contribute to the corresponding research community by making the source codes of the proposed vision framework public.
\end{enumerate}

The rest of this manuscript is organized as follows. Section II reviews conventional research outcomes relevant to the proposed vision framework and compare them. Section III describes the proposed P2CWS framework. Sections IV and V illustrate the feature extraction and pedestrian intention prediction processes in detail. Section VI delineates the evaluation settings for performance verification and the experiment results with corresponding analysis follow in Section VII. Section VIII discusses future research direction for further improvement of the proposed framework and concluding remarks follow in Section IX.

\section{Related Works}
In this section, we review previous research outcomes relevant to the proposed P2CWS framework and the two tasks. We discuss the main ideas and limitations of previous works and compare them with the proposed framework.

\subsection{Cooperative-Intelligent Transport System}

Cooperative-Intelligent Transportation Systems (C-ITS) collect and provide information in both directions between vehicles and road-side infra-structures \cite{chen2014cooperative}; in general, C-ITS utilizes on-board-unit (OBU) installed on vehicles and vision, radar and Lidar sensors on the road-side for information processing. C-ITS allows vehicles and transportation infra-structures to inter-connect, share information and coordinate pertinent actions \cite{autili2021cooperative}. Representative services of C-ITS includes slow-vehicle, pedestrian collision and abnormal condition on the road (accidents and construction) warning services. 

It is true that the initial installation of OBU and setting of transportation infra-structures for C-ITS. However, C-ITS can secure time to preemptively respond to risks that occur far from the subject vehicle and generate accurate information regarding the hazardous area---minimizing traffic accidents. Therefore, various countries such as Europe, the United States, Republic of Korea and Japan are currently conducting C-ITS demonstration projects to compensates for blind spots that drivers cannot detect by in-vehicle sensors through infrastructure sensors installed on the roadside \cite{lu2018c, lu2019pan, chen2015big}.

\subsection{Pedestrian Collision Warning Service}

Among C-ITS services, the pedestrian collision warning service aims to improve pedestrian safety at intersections or road sections. The pedestrian collision warning service implemented in the demonstration projects in various regions by applying the C-ITS standard acquires image data from a camera installed on the road-side such as crosswalks, and streams the image data to the image processing unit. The image processing unit analyzes the received image data to detect possible collisions and transmit collision warning messages to nearby vehicles through the vehicle-to-everything (V2X) server and RSU. However, due to the large amount of streaming data and the communication latency, simplification of software architecture and processing speed beyond real-time operation have become a critical concern \cite{park2019edge}. 



\subsection{Pedestrian Crossing Intention Prediction}

Prediction of pedestrian's intention is under active research to realize intelligent transportation systems and autonomous driving. Especially, it plays a key role in realizing a pedestrian collision warning service. One of the works has utilized pedestrian's demographic information, such as gender and age, and the movement of pedestrians to predict the intention \cite{zhang2020pedestrian}. Although such demographic information could help infer pedestrian's intention, pedestrian's demographic information is not always available and straightforward to recognize. 

On one hand, hand-crafted features or statistical model design still perform better than deep neural networks in certain environments \cite{rehder2014head, zhang2020pedestrian} since a sufficient amount of data for training large deep neural networks is not available in the area of the pedestrian intention prediction task. On the other hand, deep neural networks trained on large public datasets could replace sub-modules of intention prediction systems \cite{fang2019intention, neogi2020context}. Although such methods can guarantee solid performance in predefined environments, they can hardly generalize to multiple sites and they in general require a fine-tuning process to get deployed in new sites.

Another stream of research incorporates deep-learning methods to maximize the performance \cite{saleh2019contextual, song2020pedestrian, yang2018scene}. At the current state, the resulting algorithms assume specific situations such as evacuations and thus do not generalize to common transportation scenarios. Moreover, an intention prediction algorithm in signalized environments takes the signal and elapsed time of the signal phase into account in addition to environmental context, vehicle features and pedestrian characteristics \cite{gu2017human}. The work, however, does not guarantee universality over multiple sites. The work that is most relevant to our framework estimates 2D pose for predicting pedestrian intention \cite{fang2019intention}. However, the performance of intention prediction based on 2D pose degrades in the cases of occlusions and view variations.

\subsection{Human Pose Estimation}
Human pose estimation algorithms include two main categories: 2D and 3D pose estimations. Both 2D and 3D pose estimation algorithms take in RGB images and estimate the pose of humans within the images. The surge of deep-learning has resolved the limitations of classical methods in 2D human pose estimation \cite{cao2017realtime, xiao2018simple, sun2019deep}. The research on 2D pose estimation with deep-learning has become feasible with the collection of corresponding datasets \cite{lin2014microsoft, andriluka2018posetrack}. The most widely used COCO dataset contains over 200,000 images and 250,000 person instances with the labels of 17 keypoints.

The request for 3D coordinates of human joints has triggered the development of 3D pose estimation algorithms \cite{martinez2017simple, alp2018densepose, kocabas2020vibe}. Exemplary applications of 3D pose estimation encompass AR/VR, human computer interaction, computer graphics and human action understanding. The research on 3D pose estimation has become active with the collection of large datasets \cite{ionescu2014human3, mehta2017monocular, alp2018densepose}, which is similar to the case of 2D pose estimation. Recent 3D pose estimation methods have incorporated pre-trained 3D human models \cite{loper2015smpl, pishchulin2017building}. Incorporation of 3D human models improves the performance significantly and enables 3D pose estimation to overcome harsh conditions such as occlusions and view variations due to the injection of additional knowledge-base.

\section{Predictive Pedestrian Collision Warning Service}\label{sec3}

In this section, we describe the proposed predictive pedestrian collision warning service (P2CWS) framework architecture.

\subsection{Service Overview}

\begin{figure*}
    \centering
    \includegraphics[width=0.95\textwidth]{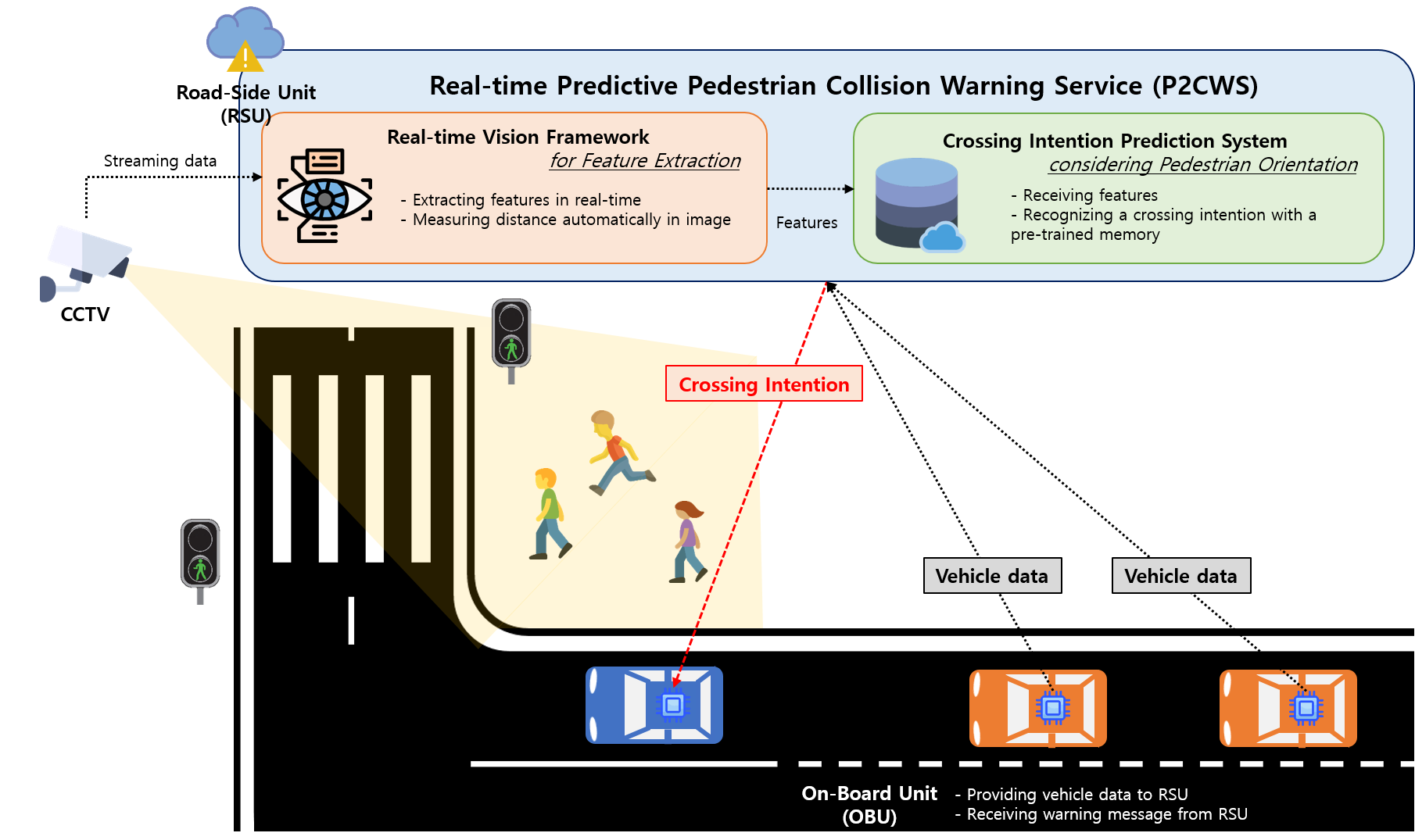}
    \caption{Overview of the proposed P2CWS framework. P2CWS resides within road-side units to collect information from pedestrians and vehicles. The framework consists of a real-time vision framework for feature extraction and a crossing intention prediction system. Moreover, P2CWS provides a real-time collision warning service to prevent possible collisions between pedestrians and vehicles.}
    \label{fig:proposed_P2CWS_overview}
\end{figure*}

We designed the P2CWS framework to alert dangerous behaviors of pedestrians to vehicle drivers approaching intersections. Fig. \ref{fig:proposed_P2CWS_overview} shows the overview of P2CWS. P2CWS require three hardware components: On-Board Unit (OBU) on vehicles, Road-Side Units (RSU), and CCTV installed at an intersection. For detecting pedestrians in real traffic scenes, we consider a camera (CCTV) as a sensor. P2CWS resides on RSU and consists of a real-time vision framework and a crossing intention prediction system. The overall data flow of P2CWS is as follows. The installed camera at an intersection shoots a fixed site including pedestrians before crossing the crosswalk, and collects a sequence of images at every 0.033 second (30 FPS). It transmits the collected image sequences to RSU. The OBU on each vehicle gather their own vehicles’ location data while uploading their locations to RSU. If RSU receives no information from OBU, the vision framework of P2CWS extracts the position and speed information of the vehicle. The real-time vision framework of P2CWS processes collected information and extracts features with which the crossing intention prediction system predicts the intention of pedestrians.

Implementing the P2CWS demands a high level of real-time data processing and transmission techniques. There are still a number of issues associated with the application of wireless cellular communication in practice, such as latency, reliability, data delivery ratio, and GPS accuracy of smartphones in vehicles. The main focus of our work is on studying a machine learning-based collision warning rather than on dealing with the communication problems---we do not consider such communication issues. Therefore, we develop the machine learning-based collision warning system under the assumption that P2CWS receives traffic information with acceptable latency and deliberated accuracy.

\subsection{Collision Warning Strategy}

\begin{figure}
    \centering
    \includegraphics[width=0.45\textwidth]{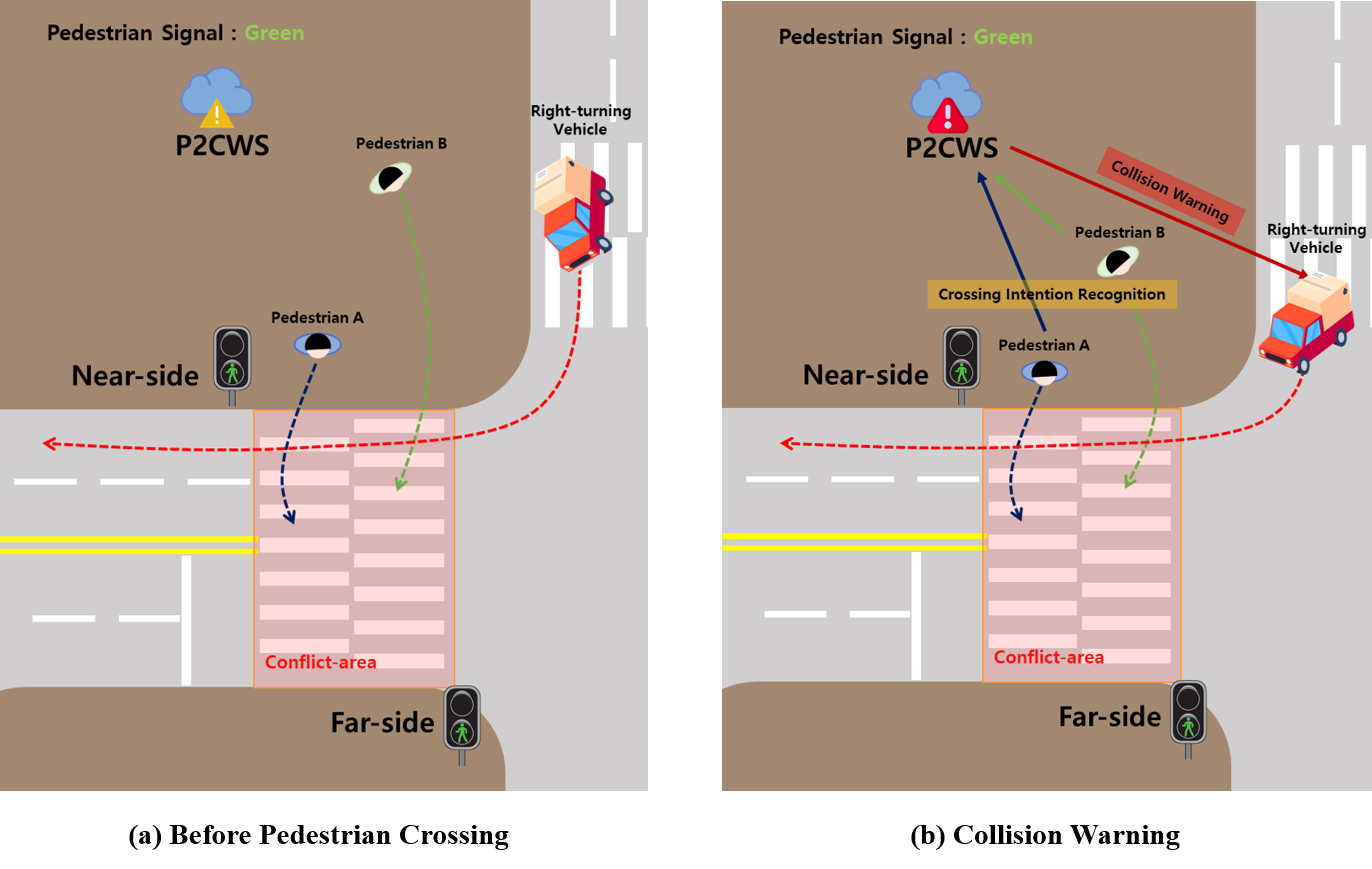}
    \caption{Collision warning strategy. The proposed P2CWS framework prevents possible accidents by predicting pedestrian's crossing-or-not-crossing intention in advance.}
    \label{fig:collision_warning_strategy}
\end{figure}

Fig. \ref{fig:collision_warning_strategy} displays the concept of collision warning provided by P2CWS. P2CWS extracts features of pedestrians, vehicles and other contexts every 0.033 seconds. Then, the crossing intention prediction system recognizes the pedestrian’s crossing intention to warn drivers approaching the intersection of possible collisions. Since the paths of vehicles are available, P2CWS could predict possible collisions. Moreover, P2CWS aims to predict the future crossing intention in 1.5 seconds. 
\section{Real-time vision framework}\label{sec4}

We describe the real-time vision framework for feature extracting in this section. The proposed real-time vision framework allows effective and efficient Prediction of crossing intention.

\subsection{System Overview}

\begin{figure*}
    \centering
    \includegraphics[width=0.95\textwidth]{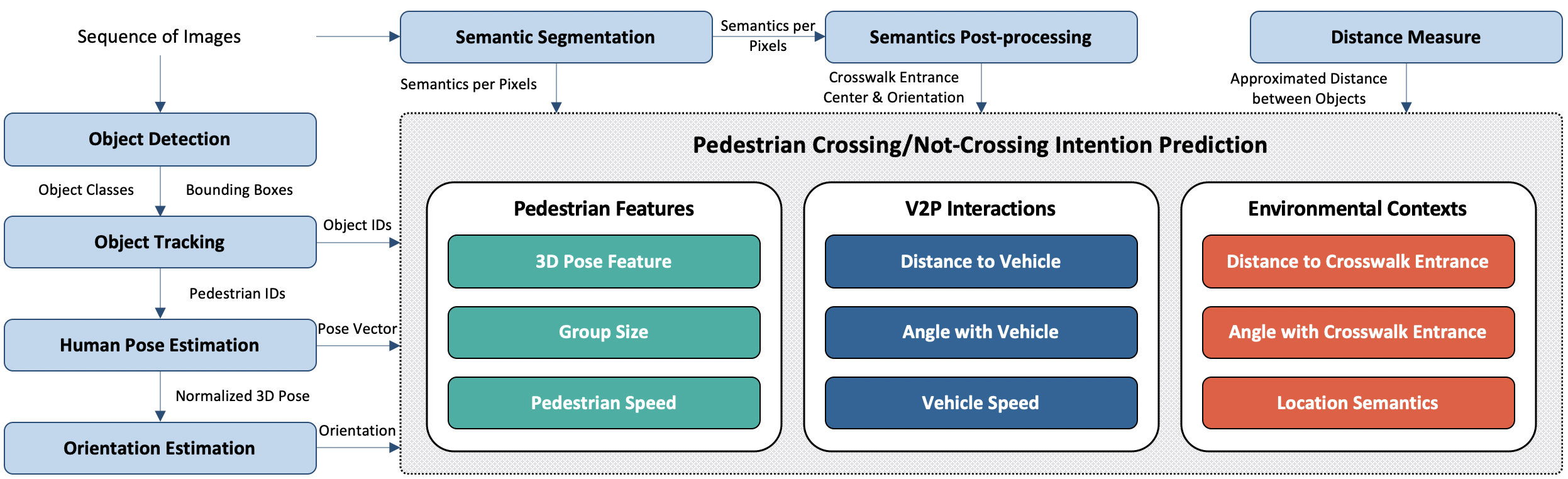}
    \caption{Overall architecture of the proposed real-time vision framework. An input sequence of images goes through a set of modules and pedestrian features and environmental contexts get extracted. Then, the process of intention prediction process follows in the crossing intention prediction system. Moreover, the proposed framework analyzes pedestrian behaviors using the extracted features.}
    \label{fig:overall_architecture}
\end{figure*}

Fig.\ref{fig:overall_architecture} illustrates the overall architecture of the proposed vision framework. The input sequence of images first passes through the object detection and object tracking modules which recognize object semantics and object identities. Then, the human pose estimation and orientation estimation modules extract pose and orientation features of pedestrians. Meanwhile, a set of first image frames goes through the semantic segmentation module for the analysis of environmental semantics. After extracting the environmental semantics, the semantic segmentation module becomes idle. Moreover, the distance measure module estimates distances between entities. After all the features necessary for the intention prediction get extracted, the process of intention prediction begins.

\subsection{Pose Estimation}

\begin{figure}
    \centering
    \subfloat[A 2D pose example]{
        \includegraphics[width=0.45\textwidth]{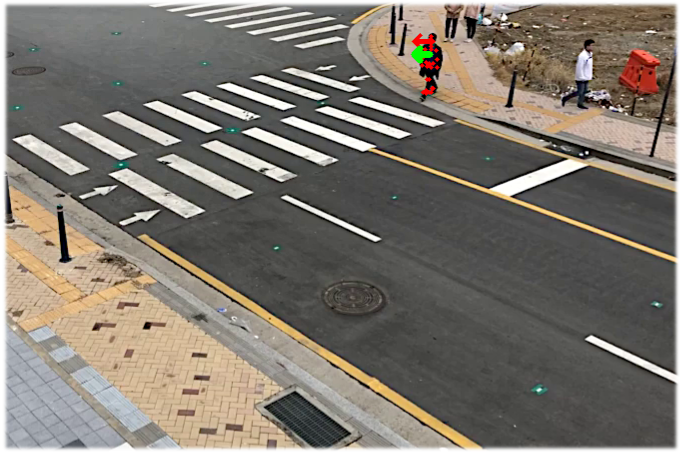}
        }\\
    \label{fig:poses_example_2d}
    ~ 
    \subfloat[A 3D pose example]{
        \includegraphics[width=0.45\textwidth]{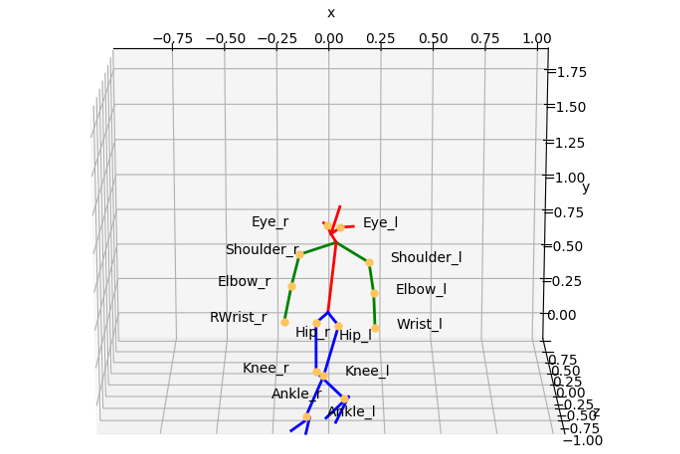}
        }
    \label{fig:poses_example_3d}
    ~ 
    \caption{A pose example. We estimated the 3D pose from the 2D pose image and projected the 3D pose into the 2D pose image (red dots). The red and green arrows in the 2D pose example indicate the head and body orientations, respectively. The orange circles in the 3D pose example represent the feature points for intention prediction.}
    \label{fig:poses_example}
\end{figure}

The proposed framework estimates both 3D and 2D poses of pedestrians (Fig. \ref{fig:poses_example}). For 3D pose estimation, the proposed framework utilizes one of the off-the-shelf 3D pose estimation algorithms \cite{kocabas2020vibe} and we propose to derive 2D poses from the estimated 3D poses.
\subsubsection{3D Pose Estimation}
The 3D pose estimation algorithm first yields the Skinned Multi-Person Linear Model (SMPL) parameters \cite{kanazawa2018end}. Then, the algorithm computes 49 joint locations in a normalized 3D space from the body vertices using a pre-trained linear regressor as follows:
\begin{equation}
    \bm{X}_{3d} = W\mathcal{M}(\theta, \: \beta),
\end{equation}
where $\mathcal{M}$ represents the SMPL model, $\theta$ and $\beta$ denote body-pose and body-shape parameters of the SMPL model, respectively and $W$ stands for the pre-trained linear regressor. We filter out redundant and non-effective joints and employ 14 joint positions for our study.

\subsubsection{2D Pose Estimation}
Once we have evaluated the 3D poses of the pedestrians in scenes, 2D pose estimation becomes a straight-forward process of projecting the estimated 3D poses into the image planes; projecting each 3D point of 3D poses generates the corresponding 2D points of 2D poses. Applying camera geometry operations projects 3D points into an image plane \cite{kim20193} as follows:
\begin{equation}
    \Vec{p}_{2d} = \frac{1}{\lambda} \cdot \bm{K} \cdot (\bm{R} \cdot \Vec{p}_{3d} + \Vec{t} \;),
\end{equation}
where $\Vec{p}_{3d} = [x, y, z]^{T}$ and $\Vec{p}_{2d} = [u, v]^{T}$ represent a point in a 3D space and the corresponding 2D point on the image plane, respectively, $\bm{R} \in \mathbb{R}^{3 \times 3}$ and $\Vec{t} \in \mathbb{R}^{3 \times 1}$ denote rotation and translation matrices, respectively, and $\bm{K} \in \mathbb{R}^{3 \times 3}$ and $\lambda$ stand for a camera intrinsic matrix and a perspective scale factor, respectively.

\subsection{Pedestrian Orientation}
We define two categories of pedestrian orientation: head orientation and body orientation. By defining two types of orientation, we can specifically analyze the orientation of pedestrians. Moreover, we use line equations in the vector form to represent orientation.

\subsubsection{Head Orientation}
We define the head orientation as the line passing through the middle point of the left and right eyes and the middle point of head-top and jaw as follows:
\begin{equation}
\begin{split}
    \Vec{v}_{1} &= t\cdot[\frac{\Vec{v}_{eye_l} + \Vec{v}_{eye_r}}{2} - \frac{\Vec{v}_{head} + \Vec{v}_{jaw}}{2}] + \frac{\Vec{v}_{eye_l} + \Vec{v}_{eye_r}}{2} \\
    &= (\frac{1+t}{2})(\Vec{v}_{eye_l} + \Vec{v}_{eye_r}) - \frac{t}{2}(\Vec{v}_{head} + \Vec{v}_{jaw}),
\end{split}
\end{equation}
where $t \in \mathbb{R}_{*}^{+} = \{x \in \mathbb{R}| x > 0\}$ is a line parameter. The example usages of the head orientation include the analysis of the pedestrian field of view.

\subsubsection{Body Orientation}
We define the body orientation as the line perpendicular to the plane containing the left and right shoulders, and mid-hip joint which passes through the middle point of the three joints as follows:
\begin{equation}
\begin{split}
    \Vec{v}_{2} = t \cdot (\Vec{v}_{shoulder_l} - \Vec{v}_{hip_m}) \times (\Vec{v}_{shoulder_r} - \Vec{v}_{hip_m}) \\
    + \frac{(\Vec{v}_{shoulder_l} + \Vec{v}_{shoulder_r} + \Vec{v}_{hip_m})}{3}.
\label{eq:body_orientation}
\end{split}
\end{equation}
The example applications of the body orientation encompass the analysis of the paths pedestrians are taking.

\subsection{Distance Measure}
Distances between objects offer a key context for the interpretation of interactions between the objects. Since 2D imaging modalities hinder the exact recovery of the 3D dimensions without prior knowledge \cite{kim2020simvodis}, we linearly approximate distances from 2D images using a knowledge-base of object dimensions as follows:
\begin{equation}
    \hat{d} = \frac{1}{2} \cdot (\frac{\Bar{h}_1}{h_1} + \frac{\Bar{h}_2}{h_2}) \cdot \sqrt{(u_1 - u_2)^2 + (v_1 - v_2)^2},
\label{eq:distance_generic}
\end{equation}
where $h$ and $\bar{h}$ denote the measured height in pixels and the mean height of an object from the knowledge-base, respectively, $(u, v)$ represents the position of an object on the 2D image plane, and $1$ and $2$ refer to object identities, respectively. Table \ref{tb:knowledge_base} displays the knowledge-base of the mean heights of the objects involved in our study. After measuring distances, we normalize them by $exp(-\hat{d}/n_{h})$ where $n_{h}$ is a normalization factor.

\begin{table}
\centering
\caption{Knowledge-base of the mean heights of objects}
\def\arraystretch{1.5}
\begin{tabular}{c || c | c | c | c | c}
\hline
\thickhline
\textbf{Object} & {Person} & {Cyclist} & {Car} & {Bus} & {Truck}\\

\thickhline
\textbf{$\bar{h}$} & $1.7$m & $1.5$m & $1.5$m & $2.5$m & $3$m\\
\hline
\thickhline
\end{tabular}
\label{tb:knowledge_base}
\end{table}
\section{Pedestrian Crossing Intention Prediction}
We illustrate the proposed intention prediction method in this section. The proposed intention prediction method consists of a feature extraction process and a classification process.

\subsection{Algorithm Overview}
\begin{figure}
    \centering
    \includegraphics[width=0.48\textwidth]{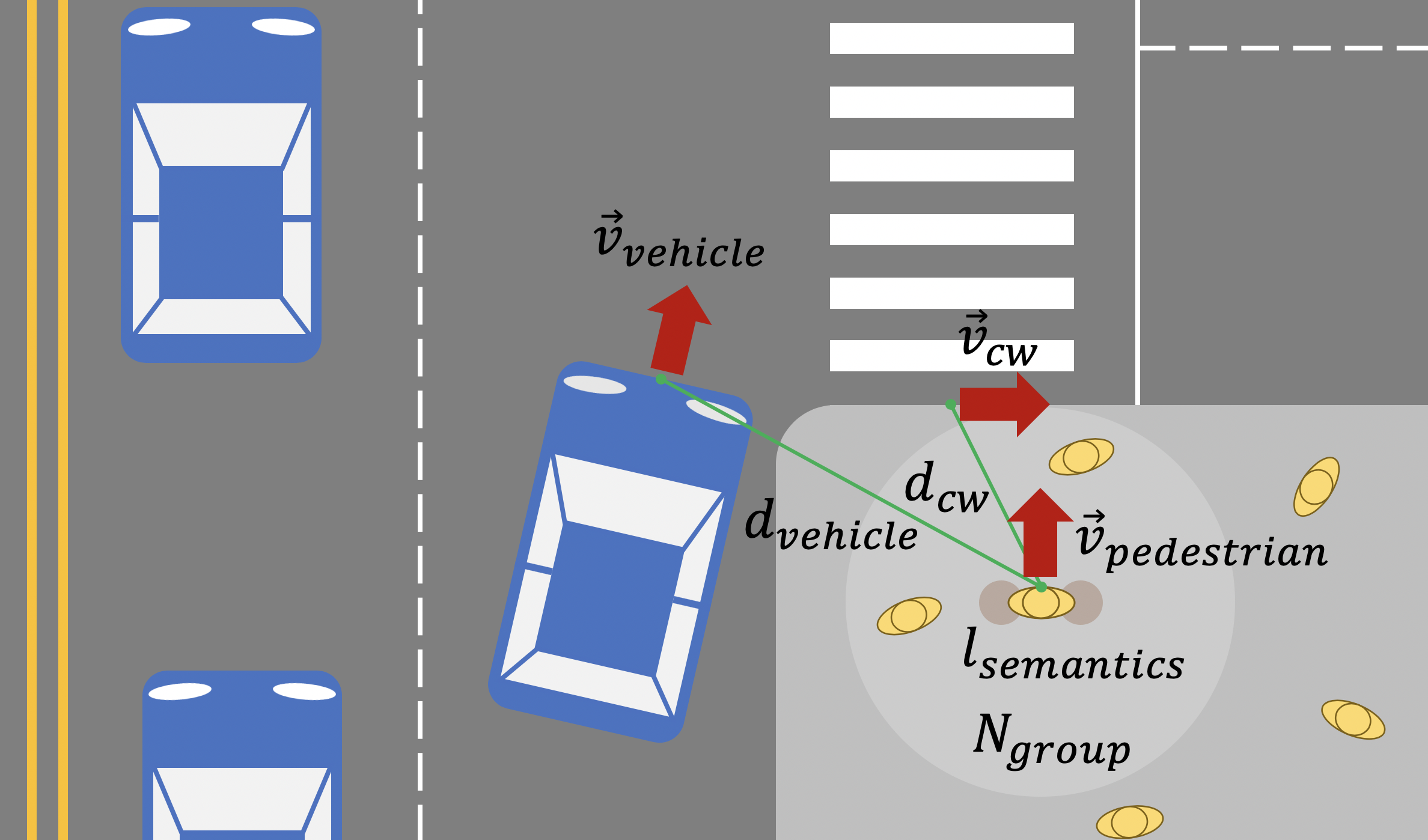}
    \caption{Overview of the features. In addition to pedestrian features, the proposed intention prediction takes the interactions between vehicles and pedestrians, and environmental contexts into account.}
    \label{fig:overview_features}
\end{figure}

Table \ref{tb:features} and Fig. \ref{fig:overview_features} summarize the features for intention prediction. We propose to extract three categories of features: pedestrian features, V2P interactions and environmental contexts. The pedestrian features derive the characteristic of pedestrians in three feature vectors (3D pose, group size and speed). The V2P interactions represent the effect of vehicles on pedestrians' intention (distance, angle and speed). Last, the environmental contexts stand for the encoding of environment information (distance to a crosswalk, angle with a crosswalk and location of pedestrian). In total, we deal with nine types of features.

\begin{table*}
\centering
\caption{List of Features for Intention Prediction}
\def\arraystretch{1.5}
\begin{tabular}{c || c | c | c | c | l}
\hline
\thickhline
\textbf{Type} & \textbf{Name} & \textbf{Notation} & \textbf{Dimension} & \textbf{Norm. Factor} & \multicolumn{1}{c}{\textbf{Description}}\\

\thickhline
\multirow{3}{*}{\textbf{\makecell{Pedestrian \\ Features}}}
 & 3D Pose    & $f_{pose}$       & $42$ & $1$  & Concatenation of fourteen 3D pose joints\\
 & Group Size & $N_{group}$      & $1$  & $10$ & Number of pedestrians in the group boundary\\
 & Speed      & $s_{pedestrian}$ & $1$  & $5$ & Moving speed of the pedestrian of interest\\
\hline
\multirow{3}{*}{\textbf{\makecell{V2P \\ Interactions}}}
 & Distance & $d_{vehicle}$  & $1$ & $10$ & Distance to the closest approaching vehicle from the pedestrian\\
 & Angle    & $\angle (\text{v2p})$ & $1$ & $1$ & Angle between the pedestrian and the vehicle\\
 & Speed    & $s_{vehicle}$  & $1$ & $10$ & Speed of the vehicle\\
\hline
\multirow{3}{*}{\textbf{\makecell{Environmental \\ Contexts}}}
 & Distance & $d_{cw}$       & $1$ & $10$ & Distance to the closest crosswalk entrance from the pedestrian\\
 & Angle    & $\angle (\text{cw})$  & $1$ & $1$ & Angle between the crosswalk entrance and the pedestrian\\
 & Location & $l_{semantic}$ & $1$ & $1$ & Pedestrian location semantics\\
\hline
\thickhline
\end{tabular}
\label{tb:features}
\end{table*}

\subsection{Pedestrian Features}
\subsubsection{3D Pose Feature}
We select 14 keypoints relevant for pedestrian movements rather than using all the keypoints extracted \cite{minguez2019pedestrian}. Fig. \ref{fig:poses_example} highlights the selected keypoints with orange circles. Other keypoints minimally vary over pedestrian movements thus offers less meaningful information. We concatenate the normalized 3D positions of 14 keypoints and form the 3D pose feature as follows:
\begin{equation}
    f_{pose} = [\Vec{v}_{arm_l}^T; \: \Vec{v}_{arm_r}^T; \: ...\:; \: \Vec{v}_{ankle_r}^T]^T.
\end{equation}

\subsubsection{Group Size}
As the size of the group that contains the pedestrian of interest affects the pedestrian decision making, we count the number of nearby pedestrians within a group boundary. We define the group boundary as a circle with 5m diameter centered at the pedestrian of interest. We normalize the group size by dividing it by $10$ before feeding it into a classifier.

\subsubsection{Speed}
We measure the speed of a pedestrian as follows:
\begin{equation}
    s_{pedestrian} = \frac{l_{t2} - l_{t1}}{\Delta t} \cdot \frac{\Bar{h}_{person}}{h_{person}},
\end{equation}
where $l_t$ is the position of the hip joint at time $t$. We track the position of the hip joint since it is the center of a body. In addition, we use $\Delta t \geq 0.5$ for a stable measurement of speed. For normalization before feeding into a classifier, we divide the measured speed by $5$.

\subsection{Vehicle-to-Pedestrian (V2P) Interactions}
We categorize vehicles into two groups: approaching or non-approaching. The distances between the pedestrian of interest and the approaching vehicles decrease ($\Delta d < 0$) over time and the distances increase ($\Delta d > 0$) in the case of non-approaching vehicles. We only consider the closest approaching vehicle for the analysis of V2P interactions. This analysis setting simplifies the analysis process and the subsequent approaching vehicles get into consideration after the closest approaching vehicle becomes a non-approaching vehicle.

\subsubsection{Distance}
The decision making of crossing or not-crossing highly depends on the distance to the approaching vehicles. Thus, we take the distance into account: the distance between the pedestrian of interest and the closest approaching vehicle. For the calculation of the distance using (\ref{eq:distance_generic}), we utilize the hip joint position of the pedestrian and the middle front position of the vehicle. We normalize the distance by dividing the measured distance by $10$.
\subsubsection{Angle}
We measure the angle between the pedestrian body orientation and the vehicle direction vector for the angle feature. The body orientation accounts for the actual direction of a pedestrian's movement and we evaluate the vehicle direction as $\Vec{v}_{vehicle} = \Vec{v}_{vehicle, t2} - \Vec{v}_{vehicle, t1}$. Since the perpendicular geometry between the pedestrian and the vehicle leads to a collision while the parallel movements of the two entities do not, we design the angle feature as follows:
\begin{equation}
    \angle (\text{v2p}) = 1 - \cos \theta = 1 - \frac{\Vec{v}_{body} \cdot \Vec{v}_{vehicle}}{|\Vec{v}_{body}| \cdot |\Vec{v}_{vehicle}|}.
\label{eq:angle_v2p}
\end{equation}
\subsubsection{Speed of Vehicle}
We measure the speed of a vehicle as follows:
\begin{equation}
    s_{vehicle} = \frac{l_{t2} - l_{t1}}{\Delta t} \cdot (\frac{2 \cdot \Bar{h}_{vehicle}}{h_{vehicle, t1} + h_{vehicle, t2}}).
\end{equation}
Since vehicles tend to move much faster than pedestrians, we compensate the scale variation by calculating the aspect ratio twice. For normalization, we apply division by $10$.

\subsection{Environment Context}
\subsubsection{Crosswalk Context}
For the crosswalk context, we calculate the distance and the angle between the pedestrian and the closest crosswalk entrance. For the distance, we consider the middle point of the crosswalk and $\Bar{h}_{person}/h_{person}$ to approximate the actual dimension from the pixel distance. For the angle, we measure $\angle (\text{cw}) = 1- \cos \theta$ as (\ref{eq:angle_v2p}). We define the direction of a crosswalk entrance ($\Vec{v}_{cw}$) with the line vector defining the crosswalk entrance.
\subsubsection{Location Semantics}
Since the current location of a pedestrian affects the crossing or not-crossing intention, we extract location semantics as one of environmental contexts. To extract semantics, we sample $8$ pixels from the nearby pixels of left and right toe joints, respectively. Among the $16$ pixels, the dominant semantic label becomes the location semantic of a pedestrian. We assign a specific number to each label to encode semantics.

\subsection{Intention Prediction}
\label{subsec:intention_prediction}
For intention prediction, the feature at time step $t$ becomes
\begin{equation}
\begin{split}
    F_{t} = [\,&f_{pose}; \;\;\;\;\: N_{group}; \: s_{pedestrian};\\
    &d_{vehicle}; \: \angle(\text{v2p}); \: s_{vehicle};\\
    &d_{cw}; \;\;\;\;\;\:\: \angle(\text{cw}); \;\; l_{semantic}\,].
\end{split}
\end{equation}
We input a set of features from a specific length of time span (temporal context) to a classifier and retrieve the intention prediction result at different future time steps. We sample 15 features per second to account for the case when detectors fail to recognize entities. Furthermore, we could attach the current state information (crossing or not-crossing) at each time step to $F_{t}$.

\section{Experimental Settings}
In this section, we delineate the experiment settings and methods for performance verification of the proposed ㅖ2ㅊㅉㄴ framework in two tasks: pedestrian orientation recognition and intention prediction tasks.

\subsection{Pedestrian Orientation Recognition}
\subsubsection{Dataset}
\begin{figure}
    \centering
    \includegraphics[width=0.45\textwidth]{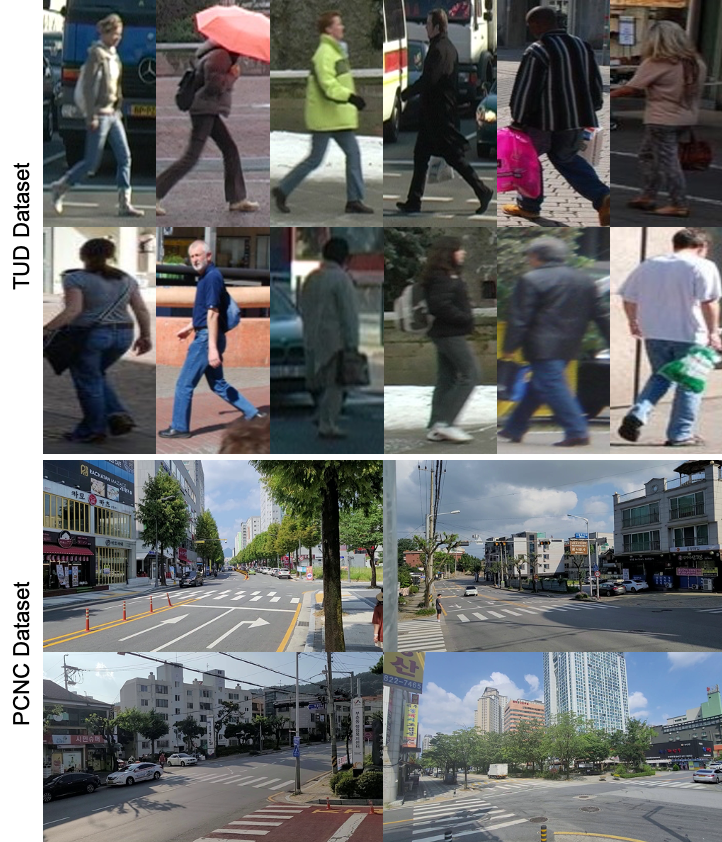}
    \caption{Data Samples. The TUD dataset includes pedestrians in various angles. The PCNC dataset collected in this work captures real-world pedestrian orientations.}
    \label{fig:data_samples}
\end{figure}

We use the TUD multi-view pedestrian dataset \cite{andriluka2010monocular} to evaluate the performance of pedestrian orientation recognition. The dataset consists of a total of 5,228 pedestrian images (refer to Fig. \ref{fig:data_samples} for sample images) and includes three subsets: training (4,732 images), validation (290 images) and test (309 images) sets. We only utilize the test set for the evaluation since the proposed vision framework functions in general cases and does not require a training step for pedestrian orientation recognition. The dataset provides the bounding boxes and the ground-truth orientation of each pedestrian ranging from $0^{\circ}$ to $360^{\circ}$ \cite{hara2017growing}.

\subsubsection{Performance Metrics}
We compute four metrics for quantitative analysis and comparison of performance: Accuracy $22.5^{\circ}$, Accuracy $45^{\circ}$, Mean Absolute Error (MAE) and Frames-per-Second (FPS). On the one hand, the accuracy metrics are defined as follows:
\begin{equation}
    \text{Accuracy} \: D^{\circ} = 100 \times \frac{1}{N_{\text{test}}} \sum_{i=0}^{N_{\text{test}}-1} \vec{\bf{1}}_{I_{D^{\circ}}}(|\theta_{i} - \hat{\theta}_{i}|),
\end{equation}
where $N_{\text{test}}$ represents the number of test images, $\theta_{i}$ and $\hat{\theta}_{i}$ denote the ground-truth and estimated orientations, respectively, $\vec{\bf{1}}_{A}(\cdot)$ is an indicator function for a set $A$, and $I_{D^{\circ}}=\{\epsilon|
\epsilon \leq D^{\circ} \}$. On the other hand, MAE is defined as follows:
\begin{equation}
    \text{MAE} = \frac{1}{N_{\text{test}}} \sum_{i=0}^{N_{\text{test}} - 1} |\theta_{i} - \hat{\theta}_{i}|.
\end{equation}

\subsubsection{Baselines}
We employ eight baseline algorithms to compare the performance of pedestrian orientation recognition (Table \ref{tb:results_orientation}). Three out of the eight involve convolutional neural networks for orientation recognition, and others involve hand-crafted features. In addition, we include the human accuracy to indicate the gap between the current state of the art and the desired performance.

\subsubsection{Implementation Details}
We compute the pedestrian orientation using (\ref{eq:body_orientation}) and do not consider the head orientation as the TUD dataset does not. To convert the pedestrian orientation vector to an angle ranging from $0^{\circ}$ to $360^{\circ}$, we measure the angle between the body orientation vector and the vector ($-1, 0, 0$) using the inner product operation. The angles increase in the clockwise direction and the angles in the third and fourth quaternions of the 2D plane range from $180^{\circ}$ to $360^{\circ}$. When calculating errors, we select the minimum values between $|\theta_1 - \theta_2|$ and $|360-\theta_1 - \theta_2|$ to account for the discontinuity between $0^{\circ}$ and $360^{\circ}$ despite their sameness.

\subsection{Intention Prediction}
\subsubsection{Dataset}
For the training and evaluation of the proposed pedestrian intention prediction algorithm, we have collected a pedestrian crossing or not-crossing (PCNC) dataset. The dataset contains 51 scenes acquired from 15 study sites and includes 64 pedestrians (refer to Fig. \ref{fig:data_samples} for sample images). We set the acquisition environment to RGB images at 30 FPS and a 1,920$\times$1,080 resolution. We have labeled the crossing and not-crossing states of pedestrians for all collected image frames.

\subsubsection{Performance Metrics}
We assess the performance of pedestrian intention prediction with two widely adopted metrics: accuracy and F1 score. The accuracy metric defined as $(TP+TN)/(TP+FP+FN+TN)$ computes the ratio of correctly predicted observation to the total observations and the F1 scores defined as $2TP/(2TP+FP+FN)$ computes the weighted average of precision and recall.

\subsubsection{Baselines}
We employ three types of classifiers and compare their performance on the intention prediction task: Feed-Forward Neural Networks (FFNN), Gated Recurrent Unit (GRU) \cite{chung2014empirical} and the encoder of transformer model \cite{vaswani2017attention}. The baseline models except for the FFNN model intrinsically entail temporal modeling. The input size at each time step is $50$ for the two temporal models, while it is $50\times T$ for FFNN where $T$ is the number of time steps as temporal context. The output size is $2$ for each time step accounting for crossing and not-crossing.

\subsubsection{Ablation Study}
First, we set the temporal context length as $0.5$ sec and let the classifiers predict the intention at the $1.5$ sec future time step. Then, we vary the model size to investigate its effect on performance. For each model we modify the number of layers ($N_{layers}$) from $2$ and $3$ to $4$ and the number of hidden units ($N_{hidden}$) from $32$ and $64$ to $128$. For the transformer model, we set the number of heads as $4$ in all cases.

Once we have found the best configuration for each model, we vary the length of the temporal context from $0.5$ sec to $3$ sec with the $0.5$ sec step size to investigate the effect of temporal context on performance. In addition, we design the classifiers to predict the intention at different future time steps (from $0.5$ sec to $2$ sec with the step size of $0.5$ sec) to examine how much classifiers can predict the future intentions. Last, we study the effect of the multi-task learning scheme \cite{ruder2017overview}, where a classifier simultaneously predicts intentions at multiple time steps.

\subsubsection{Implementation Details}
We employ Yolov3 \cite{redmon2018yolov3} as an object detector, SORT \cite{bewley2016simple} as an object tracker and HRNetV2 \cite{wang2020deep} as a semantic segmentation module. Employing other detectors and trackers did not result in dramatic performance difference. Furthermore, we split the PCNC dataset into train (50 pedestrians), validation (7 pedestrians) and test (7 pedestrians) sets. We divide the dataset by scenes rather than mixing and splitting by percentage to examine if the proposed intention prediction method could perform robustly with unseen data. We stop the training procedure when the performance on the validation set starts to decrease.

\section{Results and Analysis}
In this section, we present the experiment results in a set of different conditions, analyze the effect of various design choices, and establish the effectiveness of the proposed vision framework.

\subsection{Pedestrian Orientation Recognition}
\begin{table*}
\centering
\caption{Test Results of Pedestrian Orientation Recognition}
\def\arraystretch{1.5}
\begin{tabular}{p{40mm}|>{\centering}p{20mm}||>{\centering}p{20mm}|>{\centering}p{20mm}|>{\centering}m{15mm}|>{\centering\arraybackslash}m{15mm}}
\hline
\thickhline
\multicolumn{2}{c||}{\textbf{Method}} & \multicolumn{4}{c}{\textbf{Performance Metrics}}\\
\hline
\multicolumn{1}{c|}{\textbf{Name}} & \textbf{Features} & \textbf{Accuracy $22.5^{\circ}$} & \textbf{Accuracy $45^{\circ}$} & \multicolumn{1}{c|}{\textbf{MAE}} & \multicolumn{1}{c}{\textbf{FPS}} \\

\thickhline
Human Accuracy & - & $90.7$ & $99.3$ & $9.1$ & - \\
\hline
HOG+SVM (cost-relax) & Hand-crafted & - & $78.6$ &  - & - \\
HOG+LogReg & Hand-crafted & $57.9$ & $83.7$ &  - &-\\
CNN-based Method & CNN-based & $59.8$ & $84.5$ &  - &-\\
HOG+KRF & Hand-crafted & $62.1$ & $77.3$ & $35.2^{\circ}$ &-\\
HSSR & Hand-crafted & $66.7$ & $81.5$ &  $32.5^{\circ}$ &-\\
HOG+AKRF-VW & Hand-crafted & $68.8$ & $78.0$ &  $34.7^{\circ}$ &-\\
CNN + Mean-shift & CNN-based & $70.6$ & $86.1$ &  $26.6^{\circ}$ &- \\
Coarse-to-fine Deep Learning \cite{kim2020coarse}& CNN-based & $\textbf{72.4}$ & $89.1$ &  $\textbf{22.4}^{\circ}$  &$2.3$\\
\hline
Ours & CNN-based & $54.1$ & $\textbf{89.3}$ & $23.4^{\circ}$ & $\textbf{100.53}$ \\
\hline
\thickhline
\end{tabular}
\label{tb:results_orientation}
\end{table*}

\begin{figure}
    \centering
    \subfloat[Distribution of absolute errors]{
        \includegraphics[width=0.45\textwidth]{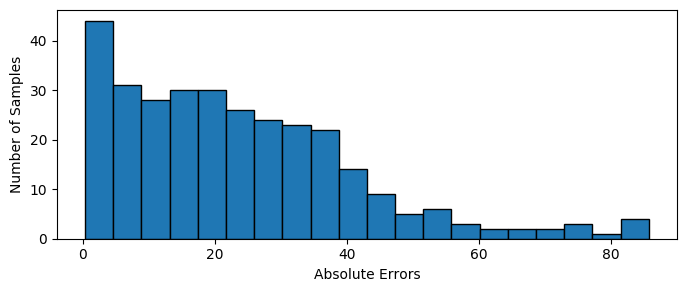}
        }\\
    \label{fig:user_behavior_scale_horizontal}
    ~ 
    \subfloat[Cumulation of absolute errors per angle]{
        \includegraphics[width=0.45\textwidth]{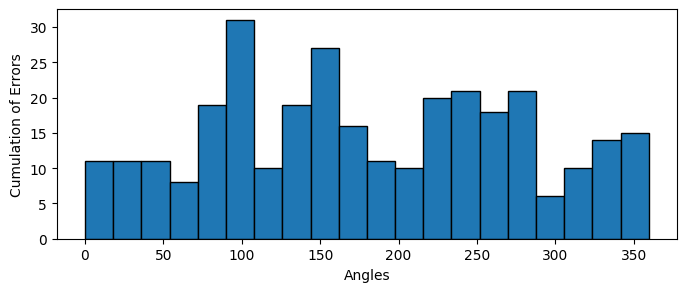}
        }
    \label{fig:user_behavior_scale_vertical}
    ~ 
    \caption{The distribution of absolute errors and cumulation of errors. The absolute errors are left-skewed, while the cumulation is relatively uniformly distributed.}
    \label{fig:result_orientation}
\end{figure}

Table \ref{tb:results_orientation} summarizes the comparative study results. The proposed vision framework outperforms or performs on par with baselines in the orientation recognition task. One thing to note is that our method does not involve any training or fine-tuning steps for orientation recognition with the TUD dataset, and we obtain the result solely with (\ref{eq:body_orientation}). This fact ensures the generality of the proposed pedestrian orientation recognition method and indicates that our method is free from the issue of overfitting. Other methods involve the process of fine-tuning for performance maximization, which could cause overfitting.

Next, our method seems to reveal weak performance in catching tiny details, although the proposed method entails much lower computational complexity (higher FPS) and displays superior performance on the overall view. The Accuracy $22.5^{\circ}$ of our method is not satisfactory compared to the performance in other metrics. We presume that the axis for measuring orientation angles might not match between the one provided by the TUD dataset and ours. Since our method measures orientations in a normalized 3D space and the TUD dataset has been annotated using 2D images, developing a calibration method for the measurement axis would result in performance enhancement in the Accuracy $22.5^{\circ}$ metric.

Moreover, our method can recognize the body orientation when pedestrians do not stand up straight while other baseline methods function with the assumption of the straight pose. The TUD dataset contains only the straight pose cases, and baseline methods trained or fine-tuned on the dataset would fail with other pedestrian postures. In a similar vein, we could easily extend our method to perform other tasks, unlike baseline methods, due to its extraction of generic 3D pose features. The example tasks our method can be extended to include pedestrian movement analysis, pedestrian action recognition, and tracking pedestrians' views.

Fig. \ref{fig:result_orientation} illustrates the distribution of absolute errors and the cumulation of absolute errors per angle. The maximum and minimum of the absolute errors are $85.70$ and $0.33$, respectively, and the number of samples decreases with the increment of the absolute error. Next, the cumulation of errors distributes uniformly over angles despite a few peaks, such as the peaks at $100^{\circ}$ and $150^{\circ}$. The uniform distribution demonstrates that our method recognizes orientations without a bias.

\subsection{Intention Prediction}

\begin{table*}
\centering
\caption{Ablation Study Result: Effect of Model Size}
\def\arraystretch{1.4}
\begin{tabular}{c || >{\centering}m{15mm} | >{\centering}m{15mm} | >{\centering}m{15mm} | >{\centering}m{15mm} | >{\centering}m{15mm} | >{\centering}m{15mm} | >{\centering\arraybackslash}m{15mm}}
\hline
\thickhline
\multicolumn{1}{c||}{\multirow{2}{*}{\makecell{\textbf{Architecture}}}} & \multicolumn{1}{c|}{\multirow{2}{*}{\makecell{\textbf{$N_{layer}$}}}} & \multicolumn{1}{c|}{\multirow{2}{*}{\makecell{\textbf{$N_{hidden}$}}}} &  \multicolumn{1}{c|}{\multirow{2}{*}{\makecell{ \textbf{$N_{param}$}}}} & \multicolumn{2}{c|}{\textbf{with State Info.}}& \multicolumn{2}{c}{\textbf{without State Info.}} \\
\cline{5-8}

& & & & \textbf{Accuracy} & \textbf{F1 Score} & \textbf{Accuracy} & \textbf{F1 Score} \\

\thickhline
\multirow{9}{*}{\textbf{FFNN}} & \multirow{3}{*}{2} & $32$ & $11,522$ & $0.9060$ & $0.9054$ & $0.5235$ & $0.4818$\\
\cline{3-8}
 & & $64$ & $23,042$ & $0.9060$ & $0.9054$ & $0.4899$ & $0.4648$\\
\cline{3-8}
 & & $128$ & $46,082$ & $0.9060$ & $0.9054$ & $0.5436$ & $0.5467$\\
\cline{2-8}
 & \multirow{3}{*}{3} & $32$ & $12,578$ & $0.9060$ & $0.9054$ & $0.5034$ & $0.4308$\\
\cline{3-8}
 & & $64$ & $27,202$ & $0.9060$ & $0.9054$ & $0.5101$ & $0.4511$\\
\cline{3-8}
 & & $128$ & $62,594$ & $0.9060$ & $0.9054$ & $0.5168$ & $0.5500$\\
\cline{2-8}
& \multirow{3}{*}{4} & $32$ & $13,634$ & $0.9060$ & $0.9054$ & $0.4899$ & $0.2830$\\
\cline{3-8}
 & & $64$ & $31,362$ & $0.9060$ & $0.9054$ & $0.5168$ & $0.4857$\\
\cline{3-8}
 & & $128$ & $79,106$ & $0.9060$ & $0.9054$ & $0.5436$ & $0.5854$\\
\hline

\multirow{9}{*}{\textbf{GRU}} & \multirow{3}{*}{2} & $32$ & $14,562$ & $0.9060$ & $0.9054$ & $0.5722$ & $0.5333$\\
\cline{3-8}
 & & $64$ & $47,554$ & $\underline{0.9128}$ & $\underline{0.9128}$ & $0.5034$ & $0.4219$\\
\cline{3-8}
 & & $128$ & $168,834$ & $\underline{0.9128}$ & $\underline{0.9128}$ & $0.5638$ & $0.5860$\\
\cline{2-8}
 & \multirow{3}{*}{3} & $32$ & $20,898$ & $\underline{0.9128}$ & $\underline{0.9128}$ & $0.5168$ & $0.2800$\\
\cline{3-8}
 & & $64$ & $72,514$ & $\underline{0.9128}$ & $\underline{0.9128}$ & $0.5503$ & $0.5180$\\
\cline{3-8}
 & & $128$ & $267,906$ & $\underline{0.9128}$ & $\underline{0.9128}$ & $\mathbf{0.6040}$ & $\mathbf{0.6380}$\\
\cline{2-8}
& \multirow{3}{*}{4} & $32$ & $27,234$ & $\underline{0.9128}$ & $\underline{0.9128}$ & $0.5839$ & $0.4918$\\
\cline{3-8}
 & & $64$ & $97,474$ & $\underline{0.9128}$ & $\underline{0.9128}$ & $0.5973$ & $0.6250$ \\
\cline{3-8}
 & & $128$ & $366,978$ & $0.9060$ & $0.9067$ & $0.5705$ & $0.6000$ \\
\hline

\multirow{9}{*}{\textbf{Transformer}} & \multirow{3}{*}{2} & $32$ & $29,394$ & $\underline{0.9128}$ & $0.9116$  & $0.5705$ & $0.4921$\\
\cline{3-8}
 & & $64$ & $36,114$ & $0.9060$ & $0.9054$  & $0.4899$ & $0.4493$\\
\cline{3-8}
 & & $128$ & $49,554$ & $0.9060$ & $0.9054$  & $0.5101$ & $0.3048$\\
\cline{2-8}
 & \multirow{3}{*}{3} & $32$ & $44,038$ & $0.9060$ & $0.9041$  & $0.5503$ & $0.4370$\\
\cline{3-8}
 & & $64$ & $54,118$ & $0.9060$ & $0.9054$  & $0.4966$ & $0.4361$\\
\cline{3-8}
 & & $128$ & $74,278$ & $0.9060$ & $0.9054$  & $0.4832$ & $0.2667$\\
\cline{2-8}
& \multirow{3}{*}{4} & $32$ & $58,682$ & $0.8859$ & $0.8811$  & $0.5570$ & $0.5147$\\
\cline{3-8}
 & & $64$ & $72,122$ & $0.9060$ & $0.9054$  & $0.4362$ & $0.3870$ \\
\cline{3-8}
 & & $128$ & $99,002$ & $0.9060$ & $0.9054$  & $0.5302$ & $0.2391$\\
\cline{2-8}

\hline
\thickhline
\end{tabular}
\label{tb:results_intention_ablation}
\end{table*}

Table \ref{tb:results_intention_ablation} summarizes the ablation study result (prediction of the intention at $1.5$ sec given $0.5$ sec context). Among the three models (FFNN, GRU and transformer), the GRU model performed the best, achieving $91.28$\% accuracy given the state information and $60.40$\% accuracy without the state information. With the state information given, all the models performed satisfactorily and the performance gaps between models were marginal. This indicates the importance of the current state information for predicting the future intention of pedestrians. The state information allowed the models to predict future intentions with a small number of model parameters.

However, there seemed to exist a performance limit that models could achieve even when the state information was given. The limit might have arisen from the data discrepancy between training and test sets. Since we had divided the dataset by the scenes rather than mixing the data and splitting by percentage, the train set would not enable models to learn a few features necessary for the perfect performance. We can overcome this limit by collecting more data, which we will conduct as future work.

Next, increasing the model size did not necessarily lead to performance enhancement. The best performing model configurations ($N_{layer}$, $N_{hidden}$) were ($4$, $128$), ($3$, $128$) and ($2$, $32$) for FFNN, GRU and transformer, respectively. At some points, models stopped improving the performance, although we increased the number of model parameters. In other words, making models wider and deeper could result in performance improvement until the performance saturation points. As we collect more data, we would be able to experiment with wider and deeper models and investigate the effect of the model size on the performance more accurately.

\begin{figure*}
    \centering
    \subfloat[FFNN Single Task]{
        \includegraphics[width=0.3\textwidth]{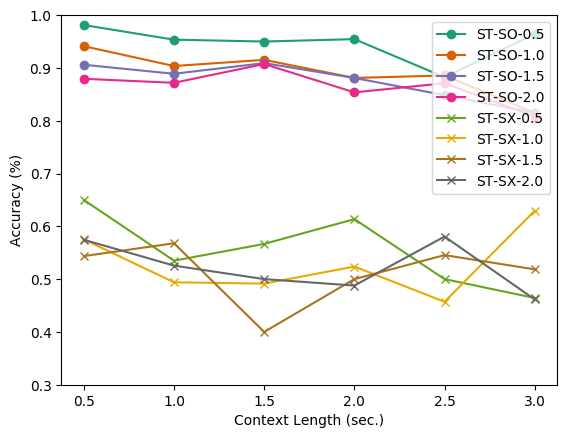}
        }
    \label{fig:intention_graph_ffnn_st}
    \subfloat[GRU Single Task]{
        \includegraphics[width=0.3\textwidth]{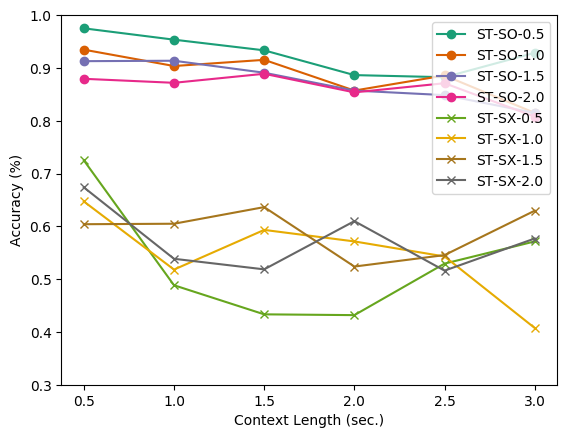}
        }
    \label{fig:intention_graph_gru_st}
    \subfloat[Transformer Single Task]{
        \includegraphics[width=0.3\textwidth]{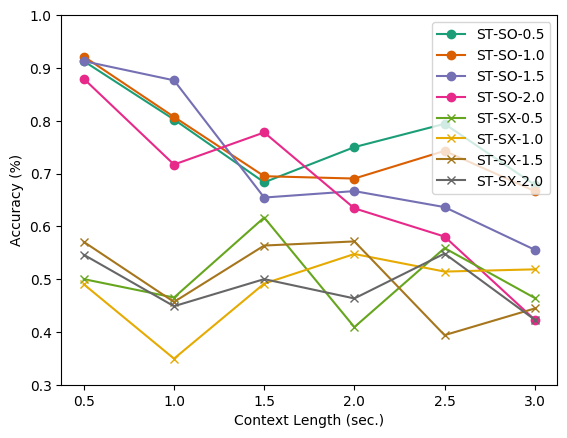}
        }\\
    \label{fig:intention_graph_transformer_st}
    \subfloat[FFNN Multi-Task]{
        \includegraphics[width=0.3\textwidth]{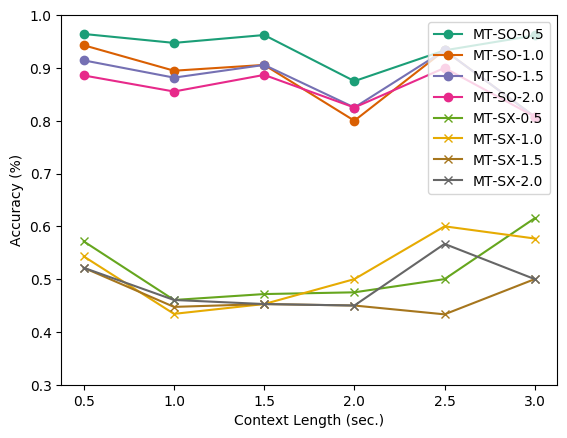}
        }
    \label{fig:intention_graph_ffnn_mt}
    \subfloat[GRU Multi-Task]{
        \includegraphics[width=0.3\textwidth]{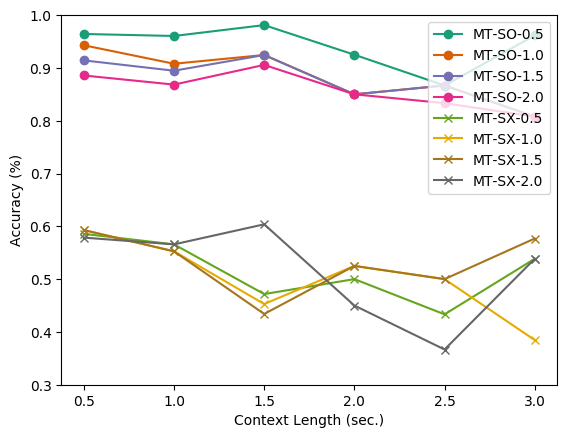}
        }
    \label{fig:intention_graph_gru_mt}
    \subfloat[Transformer Multi-task]{
        \includegraphics[width=0.3\textwidth]{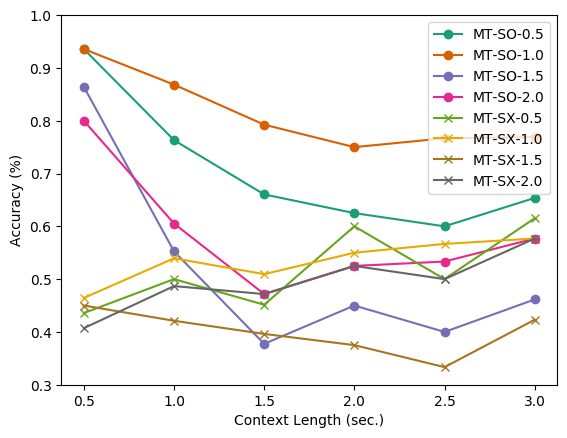}
        }
    \label{fig:intention_graph_transformer_mt}
    \caption{The effect of the context lengths. According to the given context lengths, the performance of the intention prediction at different time steps varies. Extended contexts do not persistently boost the performance. ST, MT, SO and SX in the legend denote single task, multi-task, state information given, and state information not given, respectively. The numbers in the legend represent the future time step in second.}
    \label{fig:result_intention_graph}
\end{figure*}

Fig. \ref{fig:result_intention_graph} illustrates the effect of the context lengths on the performance of predicting future intentions at various time steps. First of all, injecting more extended contexts did not improve the accuracy of the intention prediction task. In contrast, more extended contexts tended to degrade performance. We presume the short-term context plays a crucial role while the long-term context rapidly becomes obsolete in the intention prediction task. In the context of the transportation systems, the fact that the prediction of pedestrians' intention requires only short-term contexts could benefit the design of safety systems.

Furthermore, the multi-task learning setting did not consistently improve performance (the second row in Fig. \ref{fig:result_intention_graph}). In the case of GRU, the multi-task learning setting helped the models to enhance the performance with the state information given, but the performance decreased in other cases. We surmise two rationales for this phenomenon: 1) predictions of intentions at different future time steps are independent to some degree, which hinders the extraction of common features or 2) the collected data was not enough to take advantage of the multi-task learning setting.
\section{Discussion}
The proposed P2CWS framework recognizes pedestrian behaviors and predicts crossing or not-crossing intention in real-time using 3D pose estimation. Since the framework extracts site-independent features for the two tasks, it displays robust performance in multiple environments. Specifically, we demonstrated the proposed vision framework's state-of-the-art performance in pedestrian behavior recognition without any training and in intention prediction in multiple study sites. Despite its effectiveness, however, there still remain a few future works for further improvement of the performance of the proposed vision framework.

First of all, the current P2CWS framework implicitly analyzes the gaze of pedestrians, but in future studies, we can enhance the performance by explicitly analyzing the gaze. The pedestrian's gaze contains a lot of information regarding the pedestrian's future behavior. For example, whether a pedestrian has seen an oncoming vehicle determines whether a pedestrian crosses a crosswalk. Also, if a pedestrian approaches a crosswalk and frequently looks around, we can infer that the pedestrian is preparing to cross with a high probability. If an advanced vision framework tracks and analyzes the head orientation proposed in this study in real-time, it can predict pedestrians' behavior with higher accuracy.

Next, we can improve the performance by replacing individual modules with enhanced modules in the future since we designed the proposed vision framework in a modular manner. As the available data increases, the performance of deep learning-based detectors and trackers is steadily improving. Therefore, we can maximize the performance of the proposed vision framework through continuous updates. Although the currently proposed vision framework guarantees persistence and scalability to some extent, future studies can further improve the software architecture to ensure persistence and scalability more strongly. Through this architectural improvement, it will become possible to reduce the computational complexity and realize higher accuracy.

In addition, we can extend the framework to consider multiple pedestrians and multiple vehicles simultaneously. The current vision framework focuses on the intention prediction of a single pedestrian. The current vision framework can still predict the intention of multiple pedestrians through iterative analysis, but it may cause inefficiency in terms of computational complexity. In the following study, we can expand the P2CWS framework to consider multiple vehicles simultaneously and predict the intention of multiple pedestrians. This expansion will reduce the amount of computation and increase accuracy by considering interactions between entities more effectively. To this end, future research will entail the process of designing a deep neural architecture that can analyze a variable number of entities.

Last but not least, we can enhance performance through an end-to-end architecture design. Since the present P2CWS framework is modular, one can consider that the current P2CWS framework was designed through human knowledge. This design method is similar to hand-crafted features. If we replace the architecture from hand-engineered to end-to-end designs, we can expect performance advancement through feature learning. In future studies, we can maximize performance through such end-to-end designs. We can implement such end-to-end designs using image-based generative adversarial networks (GAN).
\section{Conclusion}
In this work, we proposed a real-time P2CWS framework for two central tasks in intelligent transportation systems: pedestrian orientation recognition and crossing or not-crossing intention prediction. The proposed P2CWS framework extracts site-independent features to ensure generalization over multiple sites. 3D pose estimation plays a crucial role in the feature extraction process. 3D pose estimation allows robust and accurate analysis of pedestrian orientation and crossing or not-crossing intention since it incorporates a 3D human body model as a knowledge base and temporal context within the video. We demonstrated the effectiveness of 3D pose estimation in the pedestrian behavior recognition task using the TUD dataset. The proposed pedestrian behavior recognition approach renewed the state-of-the-art performance without any finetuning process ($89.3$\% $>$ $89.1$\% in Accuracy $45^{\circ}$) and computational efficiency ($100.53$ FPS $>$ $2.3$ FPS). Furthermore, we proposed three categories of features for the intention prediction task: pedestrian features, vehicle-to-pedestrian (V2P) interactions, and environmental contexts. We verified the efficacy of the proposed approach at multiple study sites, and the proposed approach displayed $91.28$\% prediction accuracy.




\bibliographystyle{IEEEtran}
\bibliography{IEEEabrv,reference}


\newpage

\begin{IEEEbiography}[{\includegraphics[width=1in,height=1.25in,clip,keepaspectratio]{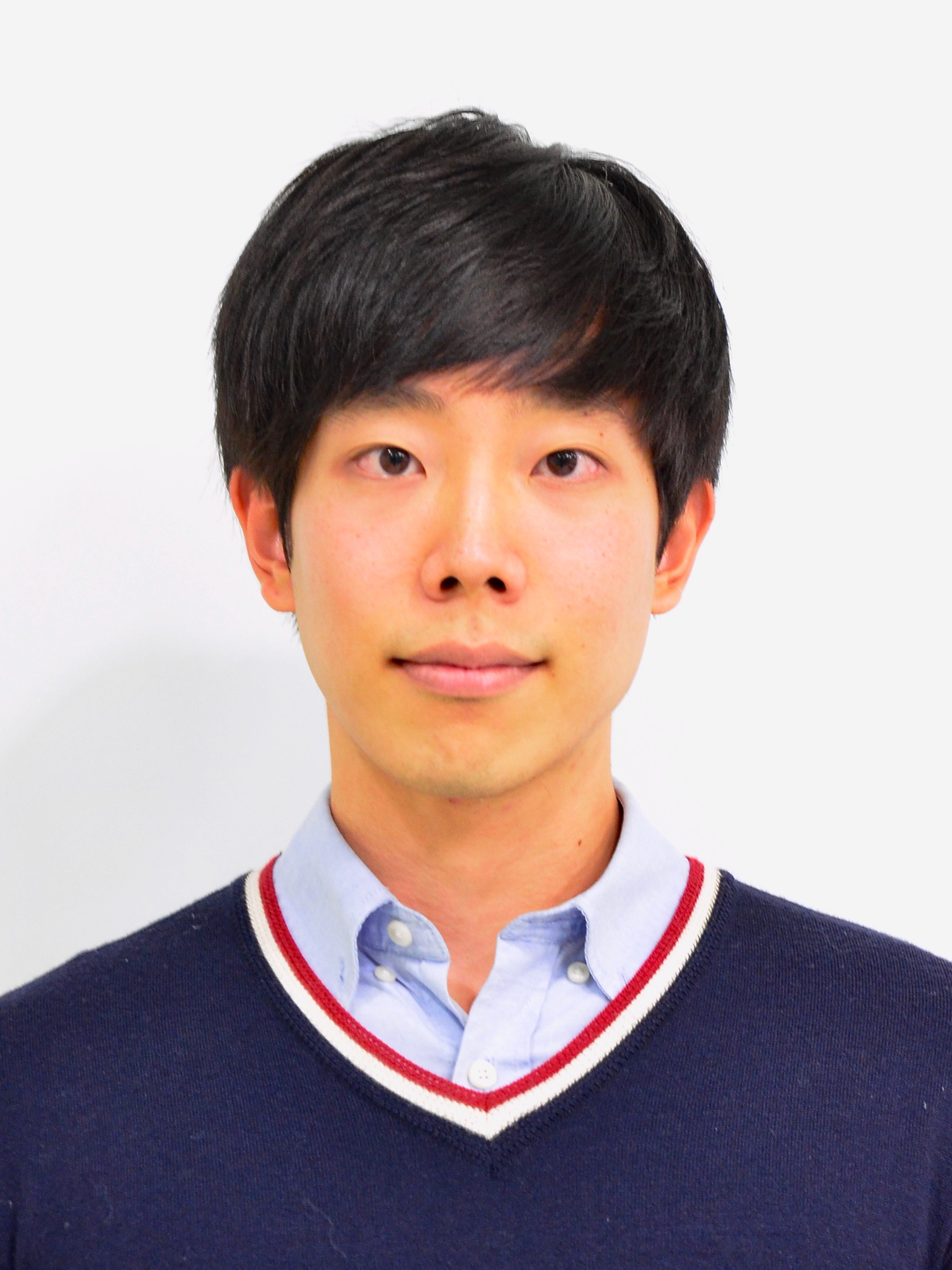}}]{Ue-Hwan Kim}
received the Ph.D, M.S. and B.S. degrees in Electrical Engineering from Korea Advanced Institute of Science and Technology (KAIST), Daejeon, Korea, in 2020, 2015 and 2013, respectively. Since 2021, he has been with the AI Graduate School, GIST, Korea, where he is leading the Autonomous Computing Systems Lab as an assistant professor. His current research interests include visual perception, service robots, intelligent transportation systems, cognitive IoT, computational memory systems, and learning algorithms.
\end{IEEEbiography}

\begin{IEEEbiography}[{\includegraphics[width=1in,height=1.25in,clip,keepaspectratio]{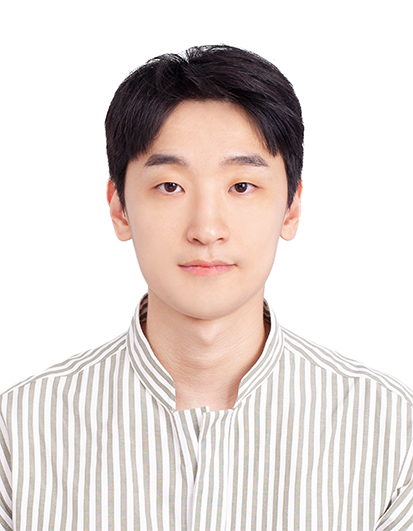}}]{Dongho Ka}
received the Ph.D degree in Civil and Environmental Engineering in 2021 and the M.S. and B.S. degrees in Industrial and Systems Engineering in 2017 and 2015 from KAIST, Daejeon, Korea, in 2017 and 2015, respectively. His current research interests include system engineering, cooperative-intelligent transport systems, AI mobility, and pedestrian behavior analysis.
\end{IEEEbiography}

\begin{IEEEbiography}[{\includegraphics[width=1in,height=1.25in,clip,keepaspectratio]{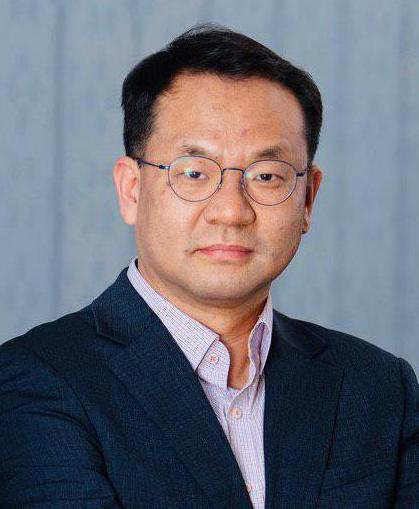}}]{Hwasoo Yeo}
received the B.S. degree in civil engineering from Seoul National University, Seoul, in 1996, and the M.S. and Ph.D. degrees in civil and environmental engineering from the University of California at Berkeley, Berkeley, CA, USA, in 2008. He is currently an Associate Professor with the Department of Civil and Environmental Engineering, KAIST. His current research interests include AI mobility, traffic flow and traffic operations, and intelligent transportation systems.
\end{IEEEbiography}

\begin{IEEEbiography}[{\includegraphics[width=1in,height=1.25in,clip,keepaspectratio]{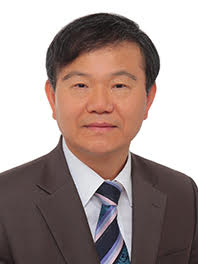}}]{Jong-Hwan Kim}
(F'09) received the Ph.D. degree in electronics engineering from Seoul National University, Korea, in 1987. Since 1988, he has been with the School of Electrical Engineering, KAIST, Korea, where he is leading the Robot Intelligence Technology Laboratory as KT Endowed Chair Professor. Dr. Kim is the Director for both of KoYoung-KAIST AI Joint Research Center and Machine Intelligence and Robotics Multi-Sponsored Research and Education Platform. His research interests include intelligence technology, machine intelligence learning, and AI robots. He has authored five books and ten edited books, and around 450 refereed papers in technical journals
and conference proceedings.
\end{IEEEbiography}

\vfill

\end{document}